\documentclass[11pt]{article}

\usepackage[final]{acl}
\usepackage{times}
\usepackage{latexsym}
\usepackage{multirow}
\usepackage{enumitem}

\usepackage[T1]{fontenc}

\usepackage[utf8]{inputenc}

\usepackage{microtype}

\usepackage{inconsolata}

\usepackage{graphicx}

\title{Bridging Lexical Ambiguity and Vision: A Mini Review on Visual Word Sense Disambiguation}

\author{Shashini Nilukshi \\
  School of Computing \\
  Informatics Institute of Technology \\
  Colombo 06, Sri Lanka \\
  \texttt{shashini.20221258@iit.ac.lk} \\\And
  Deshan Sumanathilaka \\
   School of Computing \\
  Informatics Institute of Technology \\
  Colombo 06, Sri Lanka \\
  \texttt{deshan.s@iit.ac.lk} \\ 
}

\begin{document}
\maketitle
\begin{abstract}
This paper offers a mini review of Visual Word Sense Disambiguation (VWSD), which is a multimodal extension of traditional Word Sense Disambiguation (WSD). VWSD helps tackle lexical ambiguity in vision-language tasks. While conventional WSD depends only on text and lexical resources, VWSD uses visual cues to find the right meaning of ambiguous words with minimal text input. The review looks at developments from early multimodal fusion methods to new frameworks that use contrastive models like CLIP, diffusion-based text-to-image generation, and large language model (LLM) support. Studies from 2016 to 2025 are examined to show the growth of VWSD through feature-based, graph-based, and contrastive embedding techniques. It focuses on prompt engineering, fine-tuning, and adapting to multiple languages. Quantitative results show that CLIP-based fine-tuned models and LLM-enhanced VWSD systems consistently perform better than zero-shot baselines, achieving gains of up to 6-8\% in Mean Reciprocal Rank (MRR). However, challenges still exist, such as limitations in context, model bias toward common meanings, a lack of multilingual datasets, and the need for better evaluation frameworks. The analysis highlights the growing overlap of CLIP alignment, diffusion generation, and LLM reasoning as the future path for strong, context-aware, and multilingual disambiguation systems.
\end{abstract}

\section{Introduction}

The domain of Natural Language Processing (NLP) has experienced remarkable advancements, especially with the introduction of deep learning and transformer-based architectures. These models, which can recognize intricate patterns in language through extensive data training, have significantly improved our ability to address the nuances and complications of language ambiguity \cite{yadav2021ambiguities}. Language ambiguity remains a continual challenge in NLP, as human language can convey various meanings based on the context surrounding a particular word or phrase \cite{sumanathilaka-etal-2024-llms}.

Ambiguity has historically been a significant challenge in NLP. Ambiguity can occur at various levels, primarily lexical, syntactic, and semantic, each leading to different types of interpretive uncertainty in the language \cite{Elsharif2025VisualizingAA,SUMANATHILAKA2025785}. This intrinsic ambiguity transforms many NLP tasks into fundamentally disambiguation issues. Despite advancements in NLP models, overcoming these ambiguities continues to be complicated due to factors such as insufficient context and overlapping categories of ambiguity, making linguistic ambiguity a continual area of focus in NLP research and applications \cite{OrtegaMartn2023LinguisticAA}.

\subsection{Lexical Ambiguity and Word Sense Disambiguation}
Lexical ambiguity arises when words possess several potential meanings. For instance, in the sentence “We finally reached the bank,” the term bank might refer to a “financial institution” or the “edge of a river” \cite{Li2024ATO},
Lexical ambiguity also arises when a phrase can be interpreted in several ways. For instance, the statement "I saw the man with a telescope" can imply that the man possessed the telescope or that the observer used a telescope to view him. This illustrates how the arrangement of words in a sentence can lead to multiple interpretations, which is distinct from the ambiguities found within individual words \cite{Abeysiriwardana2024ASO}.\\
Traditional Word Sense Disambiguation (WSD) aims to identify the appropriate meaning of a word based on its context from various possible interpretations. The methodologies generally fall into three primary categories: knowledge-based, supervised, and unsupervised approaches. Knowledge-based methods leverage lexical resources like dictionaries and semantic networks to derive meanings without needing labeled data \cite{bevilacqua2021recent}. Supervised techniques approach WSD as a classification problem, utilizing annotated corpora to train models that depend on contextual features. Conversely, unsupervised methods examine distributional patterns within extensive text corpora to categorize word usages by meaning without the use of labeled information. Each approach possesses distinct advantages and drawbacks, yet traditional WSD continues to be a crucial element in addressing language ambiguity in NLP \cite{sumanathilaka2024can}.

\subsection{Visual Word Sense Disambiguation}
Visual Word Sense Disambiguation (VWSD), exemplified by SemEval-2023 Task 1 \cite{Raganato2023SemEval2023T1}, addresses lexical ambiguity in image generation by selecting the correct image corresponding to the intended meaning of an ambiguous term, often with minimal textual context.
Unlike traditional WSD that relies on textual sense inventories like WordNet, VWSD requires associating a target word (e.g., "coach" in "passenger") with one correct image out of several options, demanding precise multimodal comprehension \cite{Raganato2023SemEval2023T1}. The task is challenging due to limited context, distractor images from similar domains, and the necessity of precise multimodal understanding \cite{Grbowiec2023OPIPA}.
VWSD utilizes advancements in text-to-image models (CLIP, DALL-E 2, Stable Diffusion) to connect textual and visual semantics. By resolving lexical ambiguity in a multimodal framework, VWSD enhances the semantic precision of generated images (e.g., distinguishing "bat" as an animal or equipment). This advancement is critical for applications in image retrieval, object detection, and multilingual scenarios (English, Italian, Farsi) \cite{Katyal2023teamPNAS}. The task relies on a dataset compiled from Wikidata, OmegaWiki, and BabelPic, and is evaluated using metrics such as Mean Reciprocal Rank (MRR) and Hit Rate at 1 (HIT@1) \cite{Raganato2023SemEval2023T1}.


This study makes the following key contributions:

\begin{itemize}[leftmargin=*,nosep]
    \item Provides a comprehensive comparison of Visual Word Sense Disambiguation (VWSD) methods from early feature/graph-based fusion to recent contrastive CLIP and LLM-augmented frameworks with direct performance metrics (HIT@1, MRR).
    \item Offers one of the first systematic reviews of recent advancements (2016–2025), including prompt engineering, LLM context expansion, fine-tuning, multilingual adaptation, and diffusion models, using benchmarks like SemEval-2023 Task 1.
    \item Discusses persistent VWSD challenges and proposes promising avenues for future research.
\end{itemize}
The review proceeds with a systematic literature review covering classical and modern LLM-based VWSD methods, followed by a discussion of challenges and future directions. The work encompasses all VWSD, including a thorough examination of Visual Verb Sense Disambiguation (VVSD), the foundational precursor that pioneered multimodal disambiguation techniques.

\section{Methodology}
The literature review employed a systematic approach to identify and analyze studies on VWSD, WSD, and multimodal ambiguity resolution. Searches targeted peer-reviewed papers published from 2016 to 2025 across major sources, including the ACL Anthology, arXiv, IEEE Xplore, SpringerLink, Semantic Scholar, and Google Scholar, as well as proceedings from SemEval, WACV, and CVPR. Keywords combined core and related terms such as “Visual Word Sense Disambiguation,” “Lexical Ambiguity Resolution,” “Multimodal Embedding,” “CLIP Contrastive Language-Image Pretraining,” “Visual Verb Sense Disambiguation,” and “Attention-based fusion.”.The review centers on the broader VWSD task, while also providing an in-depth analysis of VVSD efforts (2016–2022) as an important foundational predecessor.

\begin{figure}
    \centering
    \includegraphics[width=\linewidth]{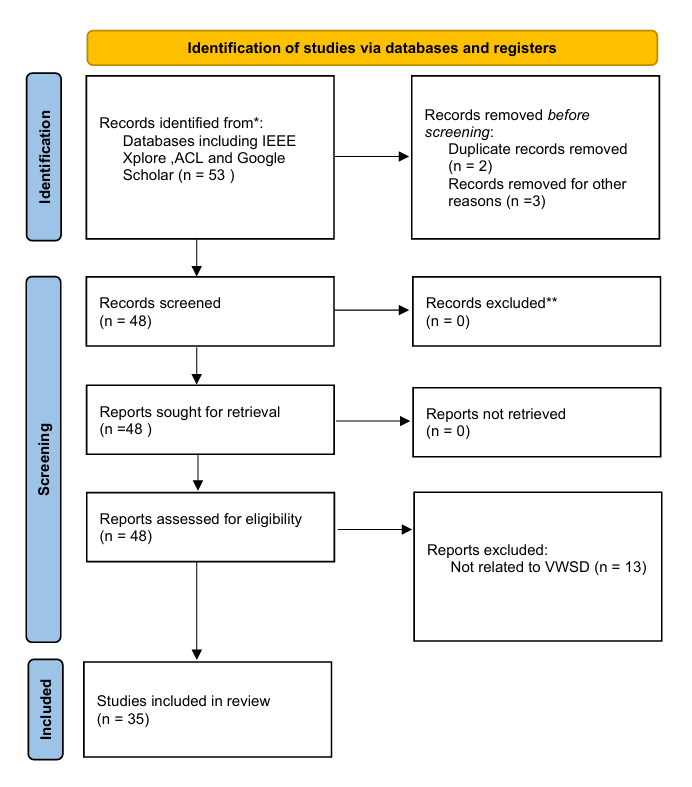}
    \caption{ PRISMA Flow of the Paper Selection Process}
    \label{fig:prisma}
\end{figure}

The selection process utilized both forward and backward citation chaining to ensure comprehensive coverage of foundational and emerging research. Inclusion criteria required studies to provide empirical results, benchmarks, or model innovations utilizing CLIP, CNNs, graph-based, or LLM-enhanced architectures, and to benchmark against SemEval-2023 Task 1 or similar multimodal datasets. Studies were excluded if they lacked empirical depth, were outside the NLP/multimodal domains, or were duplicate publications. This rigorous process ensured a representative and cutting-edge collection of literature. The paper selection and screening process is presented in Figure \ref{fig:prisma}.

\subsection{Inclusion and Exclusion Criteria}
\begin{figure*}
    \centering
    \includegraphics[width=\linewidth]{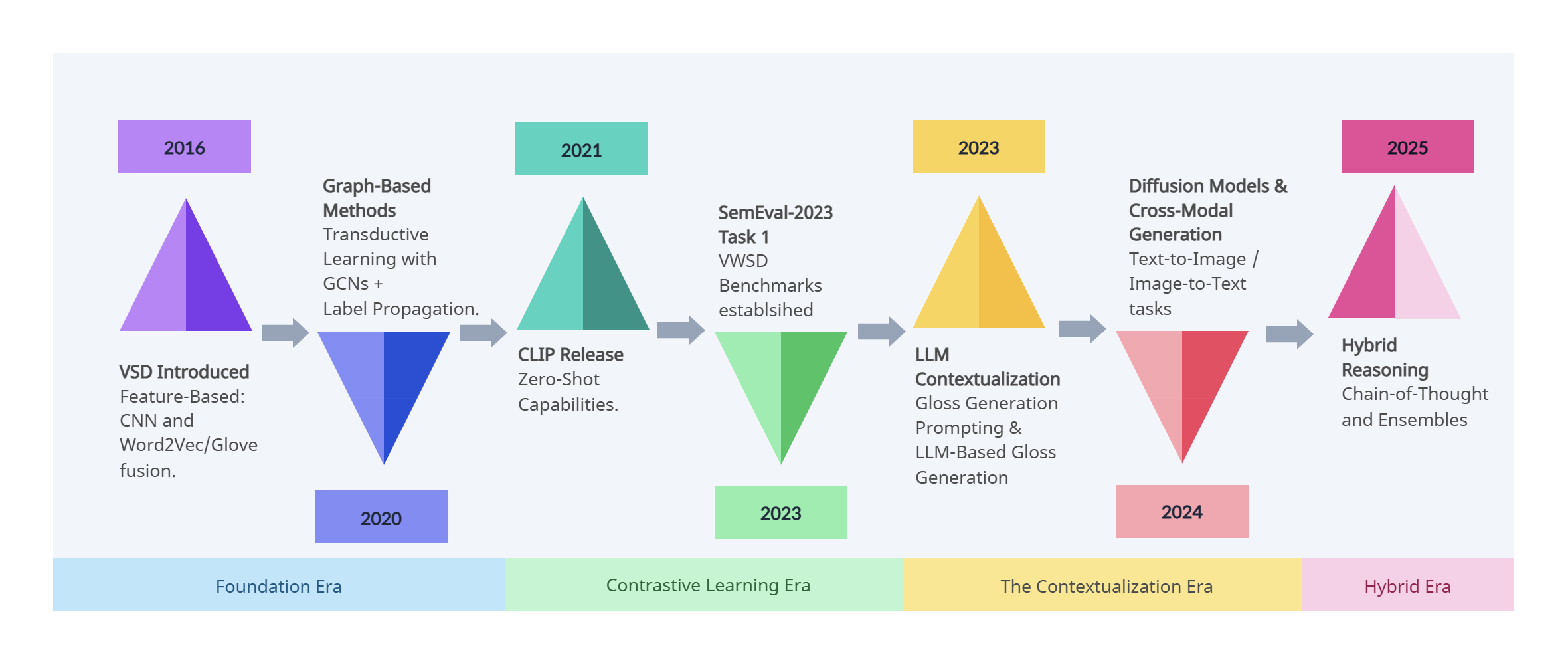}
    \caption{ Analysis of the timeline}
    \label{fig:Analysis of the timeline}
\end{figure*}

\textbf{Population/Problem (P):} We included studies focusing on Visual Word Sense Disambiguation (VWSD), Visual Verb Sense Disambiguation (VVSD), or the resolution of lexical ambiguity in multimodal contexts. Studies not concentrating on sense disambiguation or those that are purely theoretical were excluded.\\
\textbf{Intervention (I):} The review considered empirical studies utilizing innovative models such as CLIP, CNNs, graph-based techniques, or LLM-enhanced architectures, particularly those benchmarked against SemEval-2023 Task 1 or comparable datasets. We excluded papers lacking empirical substance, those presenting only theoretical models without practical implementation, and non-multimodal or solely text-based WSD approaches.\\
\textbf{Context (C):} We included peer-reviewed research articles and benchmark datasets from prominent academic platforms (e.g., ACL Anthology, CVPR, SemEval) within the fields of NLP or Multimodal AI. Excluded were studies outside these domains, duplicate publications, and papers from non-peer-reviewed or less established venues.\\
\textbf{Timeframe (T):} The review incorporates publications from 2016 to 2025 as presented in Figure \ref{fig:Analysis of the timeline}, specifically targeting advancements in multimodal fusion, contrastive learning, LLM integration, and multilingual strategies. Works published prior to 2016 were omitted.

\section{Existing Work}

\subsection{Early Multi-model Methods}
VVSD was a separate, specialized task that served as the direct predecessor to VWSD, pioneering the use of visual context to clarify verb meanings, aiming to determine the specific action illustrated in an image that corresponds to an ambiguous verb \cite{Gella2019CrosslingualVV}. For example, VVSD distinguishes between the different meanings of "play" (e.g., instrument, sport, activity) shown in an image, thereby improving the precision of multimodal applications \cite{Gella2016UnsupervisedVS}. Although VVSD initially focused exclusively on verb polysemy, the successful multimodal fusion methods developed for it motivated researchers to expand the approach to encompass all parts of speech (nouns, verbs, adjectives), leading to the creation of the more inclusive VWSD challenge. Early VVSD efforts pioneered multimodal fusion by combining traditional WSD with computer vision representations, generally classifying methods as either feature-based or graph-based \cite{Gella2016UnsupervisedVS}.

\subsubsection{Feature-Based Approaches}

\textbf{Visual Features: CNN-based Image Representations}\\
One of the pioneering techniques in VVSD involves representing images via convolutional neural networks (CNNs) such as VGGNet and ResNet. These CNNs extract hierarchical visual features that encode various levels of abstraction. For instance, \citet{Gella2016UnsupervisedVS} used CNN features from pretrained models on ImageNet to generate visual embeddings for candidate images corresponding to ambiguous words. These visual vectors serve as complementary information alongside text.

VGG (Visual Geometry Group) networks produce fixed-dimensional image feature vectors from input images capturing spatial and semantic cues \cite{Simonyan2014VeryDC}.ResNet (Residual Networks) enhanced feature extraction by enabling the training of deeper architectures with residual connections, resulting in more powerful image embeddings \cite{He2015DeepRL}.
Using these CNN-based image features enabled VVSD systems to capture fine-grained visual distinctions that are essential when the textual context is scarce or ambiguous.
\\ \\
\textbf{Textual Features: Word Embeddings and Sentence Representations} \\
On the textual side, early VVSD research utilized pretrained word embeddings like Word2Vec \cite{Mikolov2013EfficientEO} and GloVe \cite{Pennington2014GloVeGV} to represent ambiguous words and their minimal context. These embeddings encode semantic similarity based on large corpora, providing a dense vector representation of words.
Beyond simple word embeddings, sentence representations from averaging or using RNNs and LSTMs over word vectors helped capture context. However, the extremely limited context in VVSD (often one or two words around the ambiguous term) presented challenges in disambiguation solely through text.\\

\textbf{Integration of Multimodal Representations}

\begin{itemize}[leftmargin=*,nosep] 
\item Concatenation: The simplest fusion technique concatenates text and visual feature vectors into a joint representation before applying a classifier or similarity metric \cite{Gella2016UnsupervisedVS}.
\item Attention-Based Fusion: Attention mechanisms weigh visual and textual features dynamically depending on context, enabling the model to focus on salient features for disambiguation\cite{Kiela2019SupervisedMB}.
\item Canonical Correlation Analysis (CCA): CCA was employed to learn projections that maximize the correlation between visual and textual modalities into a shared latent space, improving cross-modal alignment \cite{Hardoon2004CanonicalCA,Rasiwasia2010ANA}. This method was effective in matching ambiguous words with corresponding images by ensuring similarity in the fused space.
\end{itemize}

\subsubsection{Graph Based Methods}

\textbf{Graph Construction:Building Similarity Graphs from Multimodal Features}\\
Graph-based approaches represent candidate images and senses as nodes in a graph, where edges capture similarity measures derived from both textual and visual features \cite{Vascon2020TransductiveVV}. It pioneered graph construction for VVSD by encoding multimodal similarities, enabling propagation of sense information through graph connectivity. The graph captures relationships such as: 
\begin{enumerate}[leftmargin=*,nosep] 
    \item Similarity between candidate images based on CNN visual features.
    \item Semantic similarity between word context embeddings.
    \item Multimodal similarity combining both techniques.
\end{enumerate}

This structured representation facilitates leveraging neighbourhood information and global graph properties during disambiguation\cite{Vascon2020TransductiveVV}.\\
\textbf{Label Propagation: Spreading Disambiguation Decisions}\\
Once the multimodal graph is constructed, label propagation algorithms diffuse known labels (disambiguation decisions) through the graph structure to unlabeled nodes (candidate senses with no direct annotation). This semi-supervised learning enables leveraging large amounts of unlabeled data and improves generalization. The Techniques include:

\begin{itemize}[leftmargin=*,nosep]
   \item  Random walk-based propagation.
   \item  Harmonic functions solving smoothness constraints on node labels.
   \item  Graph convolutional networks (GCNs) applied to multimodal graphs for end-to-end learning. \cite{Kipf2016SemiSupervisedCW}.\\
\end{itemize}

Transductive learning methods take advantage of both labelled and unlabeled data simultaneously to optimise performance on a fixed test set. Unlike inductive methods, they do not require generalization beyond the given dataset, which suits VVSD tasks with limited training data.
By combining feature extraction, graph structures, and label propagation, transductive learning approaches have reported substantial gains in VVSD accuracy, particularly for verbs and rare word senses. \citet{Vascon2020TransductiveVV} achieved state-of-the-art results using a transductive graph-based method on the VerSe dataset.

\subsection{Clip Based Approaches}
CLIP (Contrastive Language-Image Pretraining) has been a transformative model for VWSD, leveraging joint training on enormous image-text pairs to create a shared embedding space that aligns natural language and vision.

\subsubsection{Zero-Shot Methods}

CLIP computes cosine similarity between text and image embeddings in the shared latent space, allowing direct zero-shot disambiguation without task-specific training \cite{Radford2021LearningTV}. This enables systems to pick the image that best aligns semantically with the ambiguous word's context \cite{Poth2023MLMA}. Refining input text by crafting effective prompts can significantly improve CLIP’s performance on VWSD. For example, “A photo of a bat used in sports” versus “A bat hanging upside down” guides CLIP to focus on different senses of “bat” \cite{Ghahroodi2023SUTAS}. Recent works use prompt generation and augmentation, including synonyms and definitions to enrich minimal context \cite{Li2023RutgersMI}. And using multiple prompt templates and averaging or ensemble methods improves robustness against varied contextual expressions \cite{Taghavi2023EbhaamAS}. Diverse templates help compensate for limited information in very brief contexts.

\subsubsection{Fine-Tuned Adaptations}

VWSD-specific fine-tuning of CLIP’s joint embedding space uses contrastive losses optimized with task data, improving cross-modal alignment beyond the zero-shot baseline. SLT at SemEval-2023 Task 1 fine-tuned BLIP on VWSD datasets and showed superior mean reciprocal rank (MRR) performance \cite{Molavi2023SLTAS}. Lightweight adapters inserted into CLIP’s visual and textual encoder layers allow parameter-efficient fine-tuning. Adapters retain most pretrained weights, reducing computational costs while boosting task adaptation \cite{Poth2023MLMA}. Jointly training on related NLP and vision tasks enhances generalization for VWSD. For instance, simultaneous training with image-caption retrieval or textual WSD tasks improves shared representations \cite{Papastavrou2024ARPAAN}.

\subsection{Large Language Model Integration}

VWSD systems are increasingly utilising LLMs like GPT-3, OPT, and InstructGPT to incorporate more textual context and enhance multi-modal disambiguation. This addresses major limitations in VWSD, where phrase-level context often fails to provide sufficient semantic detail for precise image retrieval.

\subsubsection{Gloss Generation Using LLMs}
LLMs generate detailed definitions for ambiguous words, giving better input to VWSD models. \cite{Li2023RutgersMI} showed that context-aware prompts using GPT-3 perform much better than earlier systems for generating sense definitions. This is especially true for words missing from structured lexical databases like WordNet. This method, known as CADG (Context-Aware Definition Generation), increased retrieval accuracy and scalability without needing annotated datasets or additional model training \cite{Kwon2023VisionMD}.

\subsubsection{Context Expansion via LLMs}
LLMs like GPT-3 or OPT are used to turn short contexts, often just brief noun phrases, into more detailed descriptions. The \cite{Li2023RutgersMI} improved queries for VWSD by creating additional textual features, such as synonyms, hypernyms, and related concepts. This approach produced better accuracy when combined with multimodal alignment from models like CLIP and BLIP.

\subsubsection{Chain-of-Thought Reasoning with LLMs}
Chain-of-thought (CoT) prompting is becoming a useful method for step-by-step reasoning. \cite{kritharoula2023large} and related studies show that LLMs using CoT give explanations by reasoning through possible meanings, both visually and in language. They help with disambiguation choices in VWSD. This method leads to more understandable model outputs and clearer predictions.

\subsection{Cross-Modal Generation Approaches}

\subsubsection{Text to Image Generation}
Recent advances enable VWSD methods that generate modality-transformed content for disambiguation. Diffusion-based generative models synthesize images conditioned on ambiguous words with context-enhanced prompts, allowing comparison between generated and candidate images to aid disambiguation \cite{Yang2024PolCLIPAU}. Generated sense-specific images are compared quantitatively and qualitatively with candidate sets using embedding similarity and human evaluation, revealing the semantic alignment potential of generative approaches\cite{Li2023AugmentersAS}. Ensuring coherence between generated images and both the textual context and candidate images reduces ambiguity in image selection.

\subsubsection{Image to Text Generation}
 Generative caption models produce detailed textual descriptions of each candidate image, facilitating semantic matching with input context, often improving ranking and interpretability \cite{Yang2024PolCLIPAU}. Comparing generated captions to word sense glosses or expanded contexts enables integration of textual and visual features in a common semantic space. Combining multiple cues from original text, generated glosses, and image captions with weighted fusion or attention mechanisms enhances VWSD decision-making \cite{Taghavi2023EbhaamAS}. 
 
\subsection{Multilingual and Cross-Lingual Methods}

\subsubsection{Translation Based Approaches}
Non-English input contexts are translated to English for processing by predominantly English-language VWSD models, simplifying the pipeline but risking translation-induced ambiguity \cite{Patil2023RahulPA}. Models pretrained on multiple languages, including multilingual CLIP variants, encode context from different languages directly, thereby improving coverage without translation \cite{Laba2024UkrainianVW}. Using transfer learning techniques, models trained on high-resource languages are fine-tuned or adapted for low-resource language datasets, addressing data scarcity issues \cite{Yang2023TAMOS}.

\subsubsection{Language-Agnostic Methods}

More reliance on visual features and universal representations reduces dependence on language-specific context, allowing for disambiguation even with minimal or noisy text \cite{Yang2024MTAAL}. Embedding spaces are learned that align semantic content across languages and modalities, facilitating cross-lingual VWSD through language-independent features \cite{Poth2023MLMA}.
Techniques like projection or adversarial learning align text embeddings from different languages to a shared image-aligned space, enabling seamless cross-lingual VWSD \cite{Papastavrou2024ARPAAN}.

\section{Findings and Analysis}

The analysis of studies in VWSD shows a distinct progression from exclusively textual WSD to approaches that integrate multiple modalities and languages. Conventional methods for word sense disambiguation such as knowledge-based, supervised, and unsupervised offered clear interpretability but struggled in contexts with limited information \cite{dixit2015word}. The emergence of visual disambiguation reshaped the challenge as a multimodal retrieval task, harmonizing textual and visual meanings through the fusion of embeddings. Initial multi-modal systems that utilised CNN features along with Word2Vec and GloVe embeddings \cite{Gella2016UnsupervisedVS} have evolved into embedding-aligned frameworks, such as CLIP \cite{Radford2021LearningTV}, which employ contrastive learning to position images and text within shared spaces. The shift towards context-rich disambiguation is propelled by LLMS like GPT‑3 and GPT‑4, which produce glosses, synonyms, and contextually relevant expansions \cite{Patil2023RahulPA,Papastavrou2024ARPAAN}.

Recent studies \cite{Yang2024PolCLIPAU,Molavi2023SLTAS} are merging diffusion-based generators with cross-modal learning, facilitating precise sense generation and visual alignment across different languages. This progression suggests a convergence towards hybrid architectures that amalgamate CLIP, generative models (Stable Diffusion), and reasoning enhanced by LLMs, laying the groundwork for cohesive multilingual WSD systems.\\
Research results from various VWSD models reveal significant quantifiable benefits associated with strategies that incorporate multimodal and LLM enhancements. CLIP And BLIP fine-tuned approaches like ARPA \cite{Papastavrou2024ARPAAN} and SLT \cite{Molavi2023SLTAS} demonstrate better performance compared to zero-shot versions, achieving mean reciprocal rank (MRR) enhancements of up to 6--8\% over the baseline CLIP. Prompt-driven generation systems \cite{Li2023RutgersMI,Taghavi2023EbhaamAS,Ghahroodi2023SUTAS} demonstrate significant improvements in HIT@1 and consistency metrics, especially when provided with limited contextual details. Expansions of glosses based on LLM greatly improve the accuracy of WSD, the creation of definitions and the insertion of semantic glosses \cite{Li2023RutgersMI,Patil2023RahulPA} help clarify ambiguities in short-phrase contexts. Models that are multilingual and finely-tuned, like FCLL \cite{Yang2023TAMOS} and various multilingual CLIP adaptations \cite{Laba2024UkrainianVW}, uphold cross-lingual MRR by utilizing language-neutral embeddings that maintain visual semantic coherence across English, Italian, and Farsi test sets. Combination and hybrid approaches (such as ensemble deep models integrated with LLMs) reach a Mean Reciprocal Rank (MRR) of 95.77 and a HIT@1 score of 92.00, exceeding the performance of leading baseline models\cite{Setitra2025LeveragingED}. \\

This review  clearly shows that VWSD is moving towards cohesive multimodal systems that merge textual reasoning, visual grounding, and gloss-based enhancement. Contemporary architectures attain context-sensitive clarification by combining LLM-driven contextual extension with contrastive visual-text embedding approaches.

\section{Gaps and Challenges in Visual Word Sense Disambiguation}

\subsection{Context and Data Constraints }
Current benchmarks for visual word sense disambiguation are limited by a narrow range of contexts, focusing on word meanings in very specific situations. This limitation creates datasets that lack sufficient diversity and quantity, resulting in a shortage of annotated data necessary for training and evaluating effective disambiguation models. In contrast to traditional text-based WSD, where models can utilize extensive sentential or discourse context, VWSD datasets often present only minimal context typically just one or two words alongside the ambiguous target word \cite{Ogezi2023UAlbertaAS}. This severe limitation on context creates an unnaturally difficult disambiguation situation that may not accurately represent real-world scenarios where richer contextual information is generally accessible. Multiple studies have shown the important effect of this constraint: strategies for augmenting context using large language models have resulted in performance gains of 15-20 \% \cite{Li2023RutgersMI}, underscoring the essential nature of contextual information in achieving accurate disambiguation. Nevertheless, these augmentation methods also introduce additional challenges, such as the risk of hallucinations from the content generated by LLMs and the potential for error propagation throughout processing workflows \cite{Kwon2023VisionMD}. Furthermore, the field grapples with considerable data imbalance, as training datasets are heavily biased toward English and monosemous examples, whereas test sets primarily consist of polysemous instances, leading to a distribution shift that hampers model generalization \cite{Yang2023TAMOS}.

\subsection{Model Architecture Limitations}
The primary limitation in current VWSD research is the inadequacy of existing vision-language models for fine-grained sense disambiguation tasks. The leading state-of-the-art methods predominantly utilize CLIP-based architectures, which, despite their effectiveness in general multimodal applications, have several significant shortcomings regarding word sense disambiguation. Various studies have pointed out CLIP's inclination to prioritize common word meanings over less frequent or context-specific interpretations \cite{Dadas2023OPIAS}, a bias that significantly hampers disambiguation performance on polysemous terms. Moreover, these models tend to favor images that include textual representations of the target word, irrespective of their semantic relevance \cite{Dadas2023OPIAS}. The design of CLIP, which is geared towards text-image similarity instead of nuanced semantic differentiation, creates a fundamental mismatch with the demands of VWSD tasks. Furthermore, most existing systems lack end-to-end trainable architectures, often relying on zero-shot or static components that hinder task-specific tuning \cite{Li2023AugmentersAS,Taghavi2023EbhaamAS}.

\subsection{Knowledge Integration Challenges}
Current approaches to incorporating external knowledge sources into VWSD systems face significant integration challenges that limit their effectiveness. Many studies have attempted to utilise structured knowledge bases, such as WordNet, BabelNet, and Wikipedia, for definitional or contextual support; however, these integration efforts often create more issues than they resolve. WordNet-based methods, while popular, are restricted by their English-centric design and lack of visual grounding information \cite{Kwon2023VisionMD}. More importantly, using LLMs for knowledge enhancement has shown ongoing hallucination problems. These models can generate plausible but incorrect definitions, which mislead the disambiguation process \cite{Kwon2023VisionMD}. Pipeline-based architectures typically used for knowledge integration experience error propagation; mistakes in early stages, like gloss retrieval or sense matching, can cascade through the system and degrade final performance \cite{Kwon2023VisionMD}. Furthermore, the computational demands and API dependencies linked to external knowledge access create practical barriers to deployment and lead to variability in system performance\cite{Kwon2023VisionMD}.

\subsection{Multilingual and Cross-lingual Gaps}
The multilingual aspect of VWSD poses significant challenges that current research has only started to explore. A notable English bias is evident in the field. Non-English languages receive much less attention and show considerably lower performance results \cite{Ogezi2023UAlbertaAS}. This gap arises from several factors. There are not enough high-quality annotated multimodal datasets for non-English languages. Many approaches rely on translation, which adds noise and distorts meaning. Additionally, English-trained vision-language models have difficulty with cross-lingual transfer\cite{Yang2024MTAAL}. Low-resource languages encounter even greater difficulties. For instance, Ukrainian has limited coverage on Wikipedia and lacks enough visual resources for thorough disambiguation tasks \cite{Laba2024UkrainianVW}. The problem of aligning non-English textual features with shared visual representations remains largely unresolved. This issue is due to structural differences between languages and the lack of multilingual training data \cite{Ogezi2023UAlbertaAS}. These multilingual challenges significantly limit the global adoption of VWSD systems and serve as a major barrier to their wider use.

\subsection{Evaluation and Generalization Issues}
The evaluation approach in VWSD research shows important limitations in measuring the true abilities of models and their ability to generalize. Most current studies depend only on the SemEval-2023 Task 1 benchmark. This creates a limited evaluation range that may not show wider disambiguation issues or real-world performance \cite{Raganato2023SemEval2023T1}. Several studies have found notable performance drops when shifting from development to test sets, suggesting distribution shift problems that weaken claims about model generalization \cite{Berend2023SzegedAIAS}. This evaluation issue is exacerbated by the artificial nature of current benchmarks, which may not accurately reflect the complexity and variability of disambiguation situations encountered in real-world applications. Additionally, focusing on accuracy-based metrics provides little insight into how models behave in situations involving rare meanings, abstract ideas, or cross-domain transfer. The absence of standardized evaluation protocols for checking robustness, fairness across languages, or performance on out-of-distribution data creates a significant gap in the field’s ability to compare approaches meaningfully and track real progress \cite{Raganato2023SemEval2023T1}.

\section{Conclusion}
This review highlights the important progress in Visual Word Sense Disambiguation. It shows a move from traditional text-based methods to effective multi modal systems that combine vision and language. Early methods utilized handcrafted features and CNN embeddings, establishing a solid foundation. This has since evolved into contrastive learning models, such as CLIP, which link text and visual representations. Adding LLMs for context expansion and gloss generation further improves disambiguation accuracy. This helps tackle key challenges related to limited textual context. 

Multilingual and cross-modal generation methods offer promising directions for wider application across diverse languages and domains. However, current systems still have trouble with fine-grained semantic differences, a lack of context, and evaluation challenges. The combination of contrastive alignment, generative diffusion models, and LLM reasoning provides a hopeful framework for the next generation of VWSD systems. Ongoing research should focus on addressing dataset biases, expanding multilingual resources, and improving interpretability for effective real-world applications.

\section*{Limitations}
This review has several limitations that should be noted. The selection of studies was restricted to research published between 2016 and 2025, which may have left out earlier foundational work in multi modal or lexical disambiguation. The review focused on open-access publications from select repositories, such as ACL Anthology and arXiv, potentially overlooking pay walled or regional studies. Search queries used specific keywords, such as “Visual Word Sense Disambiguation” and “CLIP,” which might have missed studies that employed different terms or unpublished datasets. The analysis highlighted English-focused benchmarks like SemEval-2023, providing limited insight into non-English or low-resource contexts. Additionally, the findings primarily reflect performance metrics (HIT@1, MRR) rather than qualitative aspects such as human interpretability or cognitive plausibility. Lastly, the exclusion of ongoing or in-progress work means that the latest advances in diffusion models and large language model integration may not be fully represented.

\section*{Acknowledgments}


\bibliography{custom}

\appendix
\section{Appendices}
In this appendix, we provide an extended analysis of recent Visual Word Sense Disambiguation (VWSD) approaches. Table \ref{tab:vwsd_performance_results} offers a detailed, reference-specific breakdown of quantifiable performance benefits and techniques, while Table \ref{tab:method_comparison} provides a comparison across different VWSD method categories along with their respective core mechanisms and achieved performance metrics.

\begin{table*}
\resizebox{\textwidth}{!}{%
\begin{tabular}{|p{2.0cm}|p{2.2cm}|p{3.8cm}|p{3.5cm}|p{3.5cm}|}
\hline
\textbf{Reference} & \textbf{Core Technique} & \textbf{Specific Advancement} & \textbf{Evaluation Metrics} & \textbf{Key Challenge Addressed} \\
\hline
\citet{Papastavrou2024ARPAAN} (ARPA) & CLIP Fine-Tuning + LLM CoT & CLIP fine-tuned with Chain-of-Thought reasoning from LLMs & MRR: 6--8\% enhancement over CLIP zero-shot baseline and Accuracy over 15\% & Lack of context; interpretability of disambiguation \\
\hline
\citet{Molavi2023SLTAS} (SLT) & BLIP Fine-Tuning & Optimized contrastive loss for VWSD task-specific alignment & MRR:  67.91\% & Cross-modal alignment refinement \\
\hline
\citet{Li2023RutgersMI} & Prompt Engineering + LLM Context Augmentation (CADG) & Context-aware gloss generation using GPT-3 definitions & HIT@1: Significant improvement; Consistency metrics: Enhanced & Severe limitation of minimal textual context \\
\hline
\citet{Taghavi2023EbhaamAS} & Prompt-Driven Generation & Ensemble prompt templates with multimodal fusion & HIT@1: Significant gains; Consistency metrics: Improved & Contextual detail compensation \\
\hline
\citet{Ghahroodi2023SUTAS} & Prompt-Driven Generation & Crafted prompts guiding semantic sense selection & HIT@1: Notable improvement with limited context & Disambiguation precision with minimal input \\
\hline
\citet{Yang2023TAMOS} (FCLL) & Multilingual CLIP Fine-Tuning + Transfer Learning & Language-neutral embeddings across multiple languages & English HIT@1: 72.56\%; MRR: 82.22\% (SemEval-2023 1st place) & Cross-lingual MRR maintenance (EN, IT, FA) \\
\hline
\citet{Laba2024UkrainianVW} & Multilingual CLIP Adaptations & Cross-lingual transfer with visual semantic coherence & Cross-lingual MRR: Maintained across EN, IT, FA & Multilingual dataset scarcity \\
\hline
\citet{Patil2023RahulPA} & LLM-Based Gloss Expansion & LLM-generated definitions and semantic glosses & Accuracy: Improved WSD performance & Short-phrase ambiguity clarification \\
\hline
\citet{Setitra2025LeveragingED} & Hybrid Ensemble + LLM Integration & Multiple model outputs combined with LLM refinement & MRR: \textbf{95.77}; HIT@1: \textbf{92.00} & State-of-the-art performance via model combination \\
\hline
\textbf{CLIP Baseline} & Zero-shot CLIP & Direct cosine similarity without fine-tuning & HIT@1: 60.5\% (EN); MRR: 73.9\% (EN) & Baseline comparison reference \\
\hline
\end{tabular}%
}
\caption{Quantifiable Performance Benefits of Multimodal and LLM-Enhanced VWSD Approaches}
\label{tab:vwsd_performance_results}
\end{table*}

\begin{table*}[t] 
\label{tab:method_comparison}
\resizebox{\textwidth}{!}{%
\begin{tabular}{|p{2.5cm}|p{3.5cm}|p{4cm}|p{4cm}|}
\hline
\textbf{Method Category} & \textbf{Core Mechanism} & \textbf{Strengths} & \textbf{Evaluation (Accuracy, HIT@1, MRR)} \\ \hline
Early Multimodal & Feature or graph-based CNN/Word2Vec fusion & Combines visual (CNN) and textual (Word2Vec) features; graph-based propagation & Accuracy: up to 75.6\% (unsupervised) and 83.0\% (supervised) on the VerSe dataset (VVSD).\cite{Gella2016UnsupervisedVS}\\ \hline
CLIP-Zero Shot & Contrastive embedding alignment & No training required; leverages pre-trained vision-language alignment & Hit@1: 60.5\% (EN), 37.2\% (macro-avg); MRR: 73.9\% (EN), 54.4\% (macro-avg) [SemEval-2023 baseline] \cite{raganato-etal-2023-semeval}\\ \hline
CLIP Fine Tuned & Task-specific alignment via contrastive loss & Adapts CLIP to VWSD; fine-grained contrastive learning; parameter-efficient adapters & Hit@1: 65-72\%; MRR: 77-82\% [TAM: H@1: 72.56\%, MRR: 82.22\% \cite{yang2023tam}] \\ \hline
LLM Enriched VWSD & Gloss/context generation via GPT-3/4 & Enriches limited context; generates definitions and semantic expansions & Hit@1: 65-76\%; MRR: 73-84\% [University of Hildesheim: MRR: 73.8\% with context generation] \cite{diem-etal-2023-university} \\ \hline
Multilingual VWSD & Cross-lingual CLIP or translation-based alignment & Handles multiple languages (EN, IT, FA); language-agnostic visual representations & English Hit@1: ~84.02\%; Italian Hit@1: ~84.26\%; Farsi Hit@1: ~64\% \cite{raganato-etal-2023-semeval} \\ \hline
\end{tabular}%
}
\caption{Comparison of VWSD Methods with Performance Metrics from recent works}
\label{tab:method_comparison}
\end{table*}

\end{document}